\documentclass[5p]{elsarticle}

\usepackage{lineno,hyperref}
\usepackage{color}
\usepackage{graphicx,subfigure}
\usepackage{url}
\usepackage{amsfonts}
\usepackage{tablefootnote}
\usepackage{breqn} 
\usepackage[toc,page]{appendix}

\DeclareMathAlphabet{\mathbcal}{OMS}{cmsy}{b}{n}

\modulolinenumbers[5]

\journal{Image and Vision Computing}

\begin{document}

\begin{frontmatter}

\title{Exploiting feature representations through similarity learning, post-ranking and ranking aggregation for person re-identification}


   \author[add3,add2]{Julio C. S. Jacques Junior\corref{cor1}}
   \ead{juliojj@gmail.com}
   \author[add3,add2]{Xavier Bar\'o}
   \author[add1,add2]{Sergio Escalera}
   
   \cortext[cor1]{Corresponding author}
   \address[add3]{Faculty of Computer Science, Multimedia and Telecommunication - Universitat Oberta de Catalunya, Spain}
   \address[add2]{Computer Vision Center - Universitat Aut\`onoma de Barcelona, Spain}
   \address[add1]{Department of Mathematics and Informatics - University of Barcelona, Spain}

\begin{abstract}
Person re-identification has received special attention by the human analysis community in the last few years. 
To address the challenges in this field, many researchers have proposed different strategies, which basically exploit either cross-view invariant features or cross-view robust metrics.
In this work, we propose to exploit a post-ranking approach and combine different feature representations through ranking aggregation. Spatial information, which potentially benefits the person matching, is represented using a 2D body model, from which color and texture information are extracted and combined. We also consider background/foreground information, automatically extracted via Deep Decompositional Network, and the usage of Convolutional Neural Network (CNN) features. To describe the matching between images we use the polynomial feature map, also taking into account local and global information. The Discriminant Context Information Analysis based post-ranking approach is used to improve initial ranking lists. Finally, the Stuart ranking aggregation method is employed to combine complementary ranking lists obtained from different feature representations. Experimental results demonstrated that we improve the state-of-the-art on VIPeR and PRID450s datasets, achieving 67.21\% and 75.64\% on top-1 rank recognition rate, respectively, as well as obtaining competitive results on CUHK01 dataset.
\end{abstract}

\begin{keyword}
person re-identification, similarity learning, feature fusion, post-ranking, ranking aggregation.
\end{keyword}

\end{frontmatter}


\section{Introduction}

Person re-identification is the task of assigning the same identifier to all instances of a particular individual captured in a series of images or videos, even after the occurrence of significant gaps over time or space. It has a wide range of applications, most of them focused on surveillance and forensic systems. Even though the proposed models and reported results in this field have considerably advanced in recent years~\cite{Vezzani:2013, BedagkarGala:2014:IVC, Survey:Springer:2014}, this task still presents open challenges, mainly due to the influence of numerous real-world factors such as illumination problems, occlusions, camera settings, as well as factors associated with the dynamics of the human being, like the great variety of appearance features, pose variations and strong visual similarity between different people. These difficulties are often compounded by low resolution images or poor quality video feeds with large amounts of unrelated information, making re-identification even harder.

As related in~\cite{Cheng:CVPR:2016}, given a query person image, in order to find the correct matches among a large set of candidate images captured by different cameras, two crucial problems have to be addressed. First, good image features are required to represent both the query and the gallery images. Second, suitable distance metrics are indispensable to determine whether a gallery image contains the same individual as the query image. An ideal measurement is a matching rule that yields higher matching score for the image pairs belonging to the same person than the pairs belonging to different persons, which can be a big challenge if images are captured by different views/cameras with different setups and illumination conditions (\textit{i.e.}, a typical scenario found in person re-identification, usually not handled by direct distance metric comparison). As highlighted in~\cite{Chen:CVPR:2016}, similarity measurements which are learned (\textit{e.g.},~\cite{Yang2014, Chen:CVPR:2015}) from training samples generally enjoy better accuracy performance than learning free methods~\cite{Bazzani:CVIU:2013}. Note that the goal of metric learning algorithms is to take advantage of prior information in form of labels over simpler though more general similarity measures~\cite{KISSME:CVPR:2012}. The achieved results are then provided in the form of a list of ranked matching persons. It often happens that the true match is not ranked first but it is in the first positions. This is mostly due to the visual ambiguities shared between the true match and other ``similar'' persons~\cite{Garcia:2017}. 

In order to address the re-identification problem, existing methods exploit either feature representation~\cite{Wu:WACV:2016,Chen:IJCAI:2015,Xiao:CVPR:2016} or metric learning~\cite{KISSME:CVPR:2012, Chen:CVPR:2015}. In feature representation, robust and discriminative features are constructed such that they can be used to describe the appearance of the same individual across different camera views under various conditions~\cite{Paisitkriangkrai:CVPR:2015}, whereas distance metric learning methods attempt to learn a metric in the space defined by image features that keep features coming from same class closer, while, the features from different classes are farther apart~\cite{BedagkarGala:2014:IVC}. Recently, Convolutional Neural Networks (CNN) have been adopted in person re-identification~\cite{Li:CVPR:2014, Wu:WACV:2016}, providing a powerful and adaptive tool to handle computer vision problems without excessive usage of handcrafted image features. However, as mentioned in the work of Wu et al.~\cite{Wu:WACV:2016}, hand-crafted concatenation of different appearance features sometimes would be more distinctive and reliable, due to significant changes in view angle, lighting, background clutter and occlusion.

In this work we exploit the best of different state-of-the-art models to advance the field of person re-identification. The proposed model is inspired by the work of Chen et al.~\cite{Chen:CVPR:2016}, which enforces similarity learning with spatial constraints, and achieved (by the time of its publication) the best score (\textit{i.e.}, top rank recognition rate) on VIPeR~\cite{Gray:PETS:2007} dataset (which is one of the most challenging datasets employed in person re-identification). In this paper, by combining new and complementary features within~\cite{Chen:CVPR:2016}, followed by a post-ranking~\cite{Garcia:2017} and a ranking aggregation strategy~\cite{Prates:ICIP:2015}, we advance the state-of-the-art in person re-identification on two public datasets, VIPeR and PRID450s \cite{Roth:ACVPR:2014} (by $2.43$\% and $2.66$\%, respectively) as well as achieve competitive results on CUHK01~\cite{Li:ACCV:2012} dataset. 

The new and complementary adopted features can be briefly enumerated as follows: (i) Salient Color Names based Color Descriptor (SCNCD)~\cite{Yang2014} combined with color histogram (to encode color information), Histogram of Oriented Gradients (HOG)~\cite{Dalal:CVPR:2005} and Scale Invariant Local Ternary Patterns (SILTP)~\cite{SILTP:CVPR:2010} (to encode texture information). Although HOG and SILTP were exploited in~\cite{Chen:CVPR:2016}, they were not combined with SCNCD; (ii) SCNCD combined with background/foreground information, automatically extracted via Deep Decompositional Network (DDN) \cite{Luo:ICCV:2013}; (iii) Gaussian Of Gaussian (GOG) descriptor~\cite{Matsukawa:cvpr16}, which encodes both color and texture information; (iv) Convolutional Neural Network (CNN) features constrained by hand-crafted color his\-to\-grams \cite{Wu:WACV:2016} and combined with Local Maximal Occurrence (LOMO) features~\cite{Liao:CVPR:2015}. A quantitative analysis regarding the effectiveness of each complementary feature is presented on Sec.~\ref{complementaryfeat}. 
Experimental results showed that the proposed new features demonstrated to complement each other, being very powerful when combined with a ranking aggregation strategy.

The rest of the paper is organized as follows: Section~\ref{relatedwork} presents the state-of-the-art concerning person re-identification. The proposed model is described in Section~\ref{proposedmodel}, and experimental results are provided in Section~\ref{experimentalresults}. Finally, conclusions are given in Section~\ref{finalconsiderations}.


\section{RELATED WORK}\label{relatedwork}

Existing research on person re-identification has concentrated either on the development on sophisticated and robust features to describe the visual appearance of a person under significant visual variabilities or on the development of new learning distance metrics. In this section we present the state-of-the-art on person re-identification, briefly describing the works that achieved the best recognition rates on three broadly employed public datasets, VIPeR, PRID450s and CUHK01, without focusing on the standard taxonomy (\textit{i.e.}, feature representation or metric learning).

As in the work of Paisitkriangkrai et al.~\cite{Paisitkriangkrai:CVPR:2015}, one simple approach to exploit multiple visual features is to build an ensemble of distance functions, in which each distance function is learned using a single feature and the final distance is calculated from a weighted sum of these distance functions. However, the usage of predetermined weights is undesirable as highly discriminative features in one environment might become irrelevant in another one. In their work, a model to learn weights of these distance functions by optimizing the relative distance or by maximizing the average rank-k recognition rate is proposed. Mirmahboub et al.~\cite{Mirmahboub:2017} proposed a novel re-ranking method based on a fusion scheme that reweights an ensemble of distance metric outcomes according to their discriminative capacity. They particularly show that the fused distance perform largely better than any of the distances inferred by each feature separately.

To consider spatial information, a common usage in person re-identification is to divide the person image into few regions/stripes and concatenate dense local features to implicitly encode the spatial layout of the person. Chen et al.~\cite{Chen:CVPR:2016} proposed a model for person re-identification that combines spatial constraints and the polynomial feature map~\cite{Chen:CVPR:2015} into a unified framework. They mention that enforcing the matching within corresponding regions can effectively reduce the risk of mismatching and become more robust to partial occlusions. In addition, their framework can benefit from the complementarity of global and local similarities.

The post-ranking method for person re-identification is a relatively unexplored area~\cite{Garcia:2017} which has been  attracting a lot of attention from the research community. Prates and Schwartz~\cite{Prates:ICIP:2015} presented a Color-based Ranking Aggregation (CBRA) meth\-od, which explores different feature representations to obtain complementary ranking lists, and combine them in order to improve person re-identification. In their work, the KISSME \cite{KISSME:CVPR:2012} metric learning was adopt\-ed and different strategies for ranking aggregation, based on the Stuart rank aggregation method~\cite{Stuart:Science:2003}, were proposed. Garc\'ia et al.~\cite{Garcia:ICCV:2015,Garcia:2017} related that inspections on the ranked matches can be applied to refine the output in such a way that the correct match will have higher probability to be found in the first ranks. Hence, their work is founded on the idea that a ranking, achieved by any algorithm, contains valuable information which can be further exploited to improve the rank of the true match. To achieve such a goal, they propose an unsupervised post-ranking framework. Once the initial ranking is available, content and context sets are extracted. Then, these are exploited to remove the visual ambiguities and to obtain discriminant feature space which is finally exploited to compute the new ranking.

Bai et al.~\cite{Bai:CVPR17} studied person re-identification with man\-i\-fold-based affinity learning. In their work, a novel affinity learning algorithm called Supervised Smoothed Manifold (SSM) is proposed, which can be plunged into most existing algorithms, serving as a generic postprocessing procedure to further boost identification accuracy.

In relation to domain adaptation in machine learning, Chen et al.~\cite{Chen:IJCAI:2015} proposed a schema called Mirror Representation to address the view-specific feature distortion problem in person re-identification. It embeds the view-specific feature transformation and enables alignment of the feature distributions across disjoint views for the same person. Zhang and collaborators~\cite{Zhang:CVPR:2016} argue that most existing approaches focus on learning a fixed distance metric for all instance pairs, while ignoring the individuality of each person. They formulate person re-identification as an imbalanced classification problem and learn a classifier specifically for each pedestrian such that the matching model is highly tuned to the individual appearance. 

Considering the recently proposed CNN based methods for person re-identification, in~\cite{Wu:WACV:2016} a deep Feature Fusion Network (FFN) is proposed in order to use hand-crafted features to regularize CNN process so as to make the convolutional neural network extract features complementary to hand-crafted ones. As mentioned by the authors, different to other deep methods for person re-identification (\textit{e.g.},~\cite{Li:CVPR:2014, Ahmed:CVPR:2015}) which are based on pairwise input, they can directly extract deep features on single images, being able to be learnt by any conventional classifier. Xiao et al.~\cite{Xiao:CVPR:2016} presented a pipeline for learning deep feature representations from multiple domains with CNN. Authors argue that when training a CNN with data from all domains, some neurons learn representations shared across several domains, while some others are effective only for a specific one. Based on this observation they proposed a Domain Guided Dropout algorithm (a method of muting non-related neurons for each domain). Liu et al.~\cite{Liu:TIP17} proposed a new soft attention-based model, \textit{i.e.}, the end-to-end Comparative Attention Network (CAN), specifically tailored for the task of person re-identification, which can adaptively find multiple local regions with discriminative information in person images in a recurrent way. Such approach learns to selectively focus on parts of pairs of person images after taking a few glimpses of them and adaptively \textit{comparing} their appearances.

Although a large number of existing approaches have exploited state-of-the-art visual features, advanced metric learning algorithms, post-ranking or ranking aggregation strategies, domain adaptation based models or even CNN based ones, state-of-the-art results on commonly evaluated person re-identification benchmarks is still far from the accuracy performance needed for most real-world surveillance applications~\cite{Paisitkriangkrai:CVPR:2015}.


\section{PROPOSED MODEL}\label{proposedmodel}

\begin{figure*}[thpb]
	\centering
	\includegraphics[height=5.8cm]{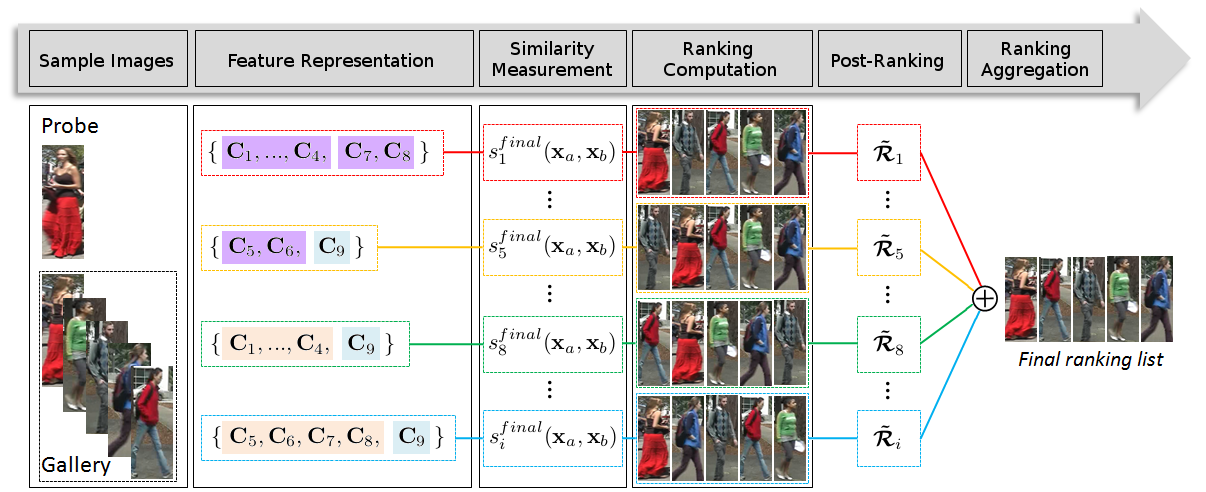}
	\caption{Overview of the proposed model. For each sample image, different visual cues are defined (\textit{i.e.}, \{$\mathbf{C}_1,...,\mathbf{C}_9$\}, as detailed in Sec.~\ref{sec:complementary-feat}). Then, different feature representations are proposed,  taking into account global (cyan regions), local (salmon regions) or both global and local (pink regions) information. For each probe image and gallery set, different similarity measures are computed using different feature representations. Each representation produces a initial ranking list based on the adopted similarity function. Then, a post-ranking approach is used in order to improve the recognition rate obtained by each initial ranking list. The final ranking list is obtained through ranking aggregation, which combines complementary ranking lists obtained from different feature representations.}
	\label{overview}
\end{figure*}

In this work, we propose to exploit different feature representations\footnote{An evaluation about different color spaces and their combinations for person re-identification can be found in~\cite{Du:ICPR:2012}.} to advance the state-of-the-art in person re-identification. In the proposed model, each image is represented in different ways, which include hand-crafted descriptors and deep features. To describe the matching between a probe image and a gallery set, a similarity learning metric built on the polynomial feature map~\cite{Chen:CVPR:2015} is adopted, also taking into account spatial (local and global) information. As each image has different descriptors, different similarities are computed, according to each representation. This way, for each probe image and gallery set, different rank lists are generated, each one assigned to each feature representation. Once these initial rankings are available, content and context information\cite{Garcia:2017} are extracted for each feature representation and respective probe image. Then, these are exploited to remove the visual ambiguities and to obtain discriminant feature space which is finally exploited to compute new ranking lists. The final rank list is obtained through ranking aggregation, which combines these complementary ranking lists. An overview of our model is illustrated in Fig.~\ref{overview}.

Next, we briefly revisit the polynomial feature map and the spatially constrained techniques~\cite{Chen:CVPR:2016}\footnote{Code available at~\url{https://github.com/dapengchen123/SCSP}}, as they are the basis of the proposed model. In a second stage, we describe the proposed complementary features. Finally, we describe the adopted post-ranking and ranking aggregation strategies.

\subsection{Polynomial Feature Map}

In order to measure the similarity between image descriptors $\mathbf{x}_a,\mathbf{x}_b \in \mathbb{R}^{d \times 1}$, we learn the similarity function as: 

\begin{equation}
f(\mathbf{x}_a,\mathbf{x}_b) = \langle \phi (\mathbf{x}_a,\mathbf{x}_b), \mathbf{W} \rangle_F,
\label{eq:similarity}
\end{equation}
where $\langle \cdot , \cdot \rangle_F$ is the Frobenius inner product. To take advantage of both Mahalanobis distance and bilinear similarity metric, we decompose $f(\mathbf{x}_a,\mathbf{x}_b)$ as follows:

\begin{dmath}
f(\mathbf{x}_a,\mathbf{x}_b) = \langle \phi_M (\mathbf{x}_a,\mathbf{x}_b), \mathbf{W}_M \rangle_F + \langle \phi_B (\mathbf{x}_a,\mathbf{x}_b), \mathbf{W}_B \rangle_F.
\label{eq:similarity:decomposed}
\end{dmath}
The part $\langle \phi_M (\mathbf{x}_a,\mathbf{x}_b), \mathbf{W}_M \rangle_F = (\mathbf{x}_a - \mathbf{x}_b)^\top \mathbf{W}_M (\mathbf{x}_a - \mathbf{x}_b)$ is connected to the Mahalanobis distance. The part $\langle \phi_B (\mathbf{x}_a,\mathbf{x}_b), \mathbf{W}_B \rangle_F = \mathbf{x}_a^\top \mathbf{W}_B \mathbf{x}_b + \mathbf{x}_b^\top \mathbf{W}_B \mathbf{x}_a$ corresponds to bilinear similarity. Both parts ensure the effectiveness of $f(\mathbf{x}_a,\mathbf{x}_b)$. 
The dimensionality of the feature map is reduced by means PCA for $\mathbf{x}_a$ and $\mathbf{x}_b$ before its generation\footnote{A detailed explanation about how $\mathbf{W}_M$ and $\mathbf{W}_B$ are learned using the ADMM optimization algorithm can be found in~\cite{Chen:CVPR:2016}.}.
\subsection{Spatially Constrained Similarity Function}

\subsubsection{Regional feature map} The input image is partitioned into $R$ non-overlap horizontal stripe regions. Each region is divided into a collection of overlapped patches, from which we extract color and texture histograms. The extracted histograms belonging to a same stripe region are concatenated together. After that, PCA is applied to reduce the dimensionality and to obtain the region descriptor $\mathbf{x}^r$ for the $r$-th stripe, where $r \in \{1,...,R\}$. A stripe region $r$ can be described by $C$ visual cues \{$\mathbf{x}^{r,1},..., \mathbf{x}^{r,c},...,\mathbf{x}^{r,C}$\}, thus $\mathbf{x}_a$ and $\mathbf{x}_b$ accordingly form $C$ polynomial feature maps for the $r$-th region, \textit{i.e.}, \{$\phi^{r,1}(\mathbf{x}_a,\mathbf{x}_b),..., \phi^{r,c}(\mathbf{x}_a,\mathbf{x}_b),..., \phi^{r,C}(\mathbf{x}_a,\mathbf{x}_b)$\}, where $\phi^{r,c}(\mathbf{x}_a,\mathbf{x}_b) = \phi(\mathbf{x}_a^{r,c},\mathbf{x}_b^{r,c})$.

\subsubsection{Local similarity integration} In order to exploit the complementary strengths of multiple visual cues within a local region, a linear similarity function is employed to combine them together for the $r$-th region:

\begin{equation}
s^r(\mathbf{x}_a,\mathbf{x}_b) = \sum_{c=1}^{C}{\langle \phi^{r,c} (\mathbf{x}_a,\mathbf{x}_b), \mathbf{W}^{r,c} \rangle_F },
\end{equation}
where $\mathbf{W}^{r,c}$ = $[\mathbf{W}_M^{r,c}, \mathbf{W}_B^{r,c}]$ and $\mathbf{W}_M^{r,c}$, $\mathbf{W}_B^{r,c}$ correspond to $\phi_M^{r,c} (\mathbf{x}_a,\mathbf{x}_b)$ and $\phi_B^{r,c} (\mathbf{x}_a,\mathbf{x}_b)$, respectively. The local similarities scores are integrated as:

\begin{equation}
s^{local}(\mathbf{x}_a,\mathbf{x}_b) = \sum_{r=1}^{R}{s^r (\mathbf{x}_a,\mathbf{x}_b)}.
\end{equation}

\subsubsection{Global-local collaboration} In order to describe the matching of large patterns across the stripes, the polynomial feature map is also used for the whole image, yielding global similarity:

\begin{equation}
s^{global}(\mathbf{x}_a,\mathbf{x}_b) = \sum_{c=1}^{C}{\langle \phi^{G,c} (\mathbf{x}_a,\mathbf{x}_b), \mathbf{W}^{G,c} \rangle_F },
\end{equation}
%
%
\sloppy where $\mathbf{W}^{G,c}$ = $[\mathbf{W}_M^{G,c}, \mathbf{W}_B^{G,c}]$ and $\mathbf{W}_M^{G,c}$, $\mathbf{W}_B^{G,c}$ correspond to $\phi_M^{G,c} (\mathbf{x}_a,\mathbf{x}_b)$ and $\phi_B^{G,c} (\mathbf{x}_a,\mathbf{x}_b)$, respectively. Here, $\phi^{G,c}(\mathbf{x}_a,\mathbf{x}_b) = \phi (\mathbf{x}_a^{G,c},\mathbf{x}_b^{G,c})$ and $\mathbf{x}_a^{G,c}, \mathbf{x}_b^{G,c}$ are the $c$-th type global visual descriptors for image $a$ and $b$. Finally, the global similarity and local similarity are linearly combined, and the overall similarity score is given by:

\begin{equation}
s^{final}(\mathbf{x}_a,\mathbf{x}_b) = s^{local}(\mathbf{x}_a,\mathbf{x}_b) + \gamma s^{global}(\mathbf{x}_a,\mathbf{x}_b),
\label{finalsimilarity}
\end{equation}
where $\gamma$ is the hyper-parameter that mediates the local and global similarities (experimentally set to $\gamma=1.1$).

\subsubsection{Visual Cues and Parameter settings} In the original model of~\cite{Chen:CVPR:2016}, four visual cues are used (\textit{i.e.}, $C=4$). First, images are resized to 48$\times$128. Each region $r$ (from $R$, experimentally set to $R=4$)\footnote{A default $R$ value was adopted from~\cite{Chen:CVPR:2016}.} is divided into a set of local patches (with 8$\times$16 of size and stride of 4$\times$8). For each patch, six types of features are extracted: HSV$_1$, LAB$_1$ (are 8$\times$8$\times$8 joint histograms), HSV$_2$, LAB$_2$ (are 48 bin concatenated histograms with each channel having 16 bins), HOG~\cite{Dalal:CVPR:2005} and SILTP~\cite{SILTP:CVPR:2010} (texture descriptors). The four visual cues $\mathbf{C}_1$, $\mathbf{C}_2$, $\mathbf{C}_3$ and $\mathbf{C}_4$ concatenate both color and texture features, which are organized as HSV$_1$/HOG, HSV$_2$/SILTP, LAB$_1$/SILTP and LAB$_2$/HOG, respectively. 

Regarding each visual cue, descriptors generated for each patch, within a specific region $r$, are concatenated to compose the descriptor of such region. Similarly, the global descriptor is generated through the concatenation of the descriptors computed for all patches. For each visual cue, obtained color and texture descriptors are normalized (to have unit $L_2$ norm) before concatenation. Then, each visual cue is reduced by PCA. Finally, the resulting descriptor is normalized again in the same way. As mentioned in~\cite{Chen:CVPR:2016}, the PCA reduced dimension $d$ depends on the size of training data. In our experiments we adopted $d$ to be $120$ for all evaluated datasets.

\subsection{Complementary features}\label{sec:complementary-feat}

In order to improve state-of-the-art recognition performance in person re-identification, we propose to include new and complementary features within the similarity function presented in~\cite{Chen:CVPR:2016}, as described next.

\subsubsection{SCNCD~\cite{Yang2014}}\label{scncdfeat} 
For each color to be named, salient color names indicate that a color only has a certain probability of being assigned to several nearest color names, and that the closer the color name is to the color, the higher probability the color has of being assigned to this color name. Through this way, we can assign multiple similar colors to the same index with the same color descriptor.

\begin{figure}[thpb]
	\centering
	\includegraphics[height=3cm]{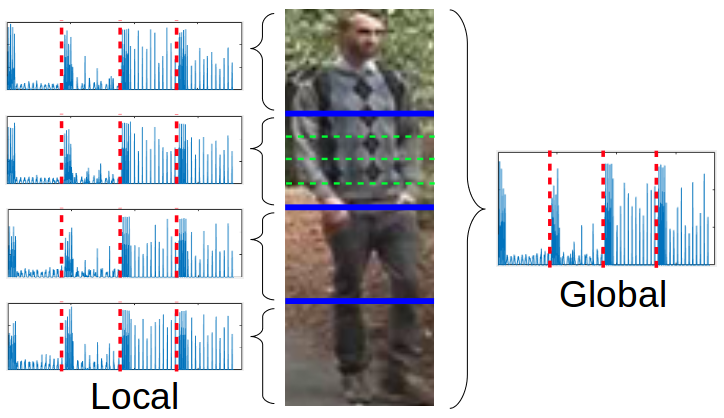}
	\caption{Illustration of the generated global and local descriptors based on SCNCD (and color histograms).}
	\label{scncdcolorhist}
\end{figure}

Color distributions over color names in different color spaces are then obtained and fused to generate a feature representation. In addition, and similarly to~\cite{Yang2014}, color histogram is computed for each color channel and fused with color names distribution (the number of bins is set to $32$). In this work, SCNCD and color histograms are extracted using the original RGB, normalized \textit{rgb}, $l_1l_2l_3$ and HSV color models. Such procedure is performed locally, regarding each region $r$, as well as globally, regarding the whole image. To be specific, SCNCD are extracted similarly to~\cite{Yang2014}, except that in our model the image is divided in 4 regions ($R=4$) and a second subdivision at the local level is performed. First, the image is divided into $R$ horizontal stripes, from which features are extracted and concatenated (global descriptor), as illustrated on the right side of Figure~\ref{scncdcolorhist}. Then, each region $r$ is subdivided again into $R$ horizontal stripes, from which features are extracted and concatenated (local descriptor), as illustrated on the left side of Figure~\ref{scncdcolorhist}.

Two new visual cues are then proposed, $\mathbf{C}_5$ and $\mathbf{C}_6$. Both concatenate color and texture features, which are organized as SCNCD/HOG and SCNCD/SILTP, respectively. In this case, HOG and SILTP are extracted in the same way as in~\cite{Chen:CVPR:2016}. As before, obtained descriptors that compose each new visual cue are normalized (to have unit $L_2$ norm) before final concatenation. The resulting descriptor is then reduced by means of PCA before final normalization step.

\subsubsection{Background/foreground information}\label{contextinformation} Due to the fact that the background in person re-identification is not constant and may even include disturbing factors, background feature representation combined directly with the foreground feature representation may reduce classification accuracy. To address this problem, \cite{Yang2014} proposed an image-foreground feature representation, which can be seen as that the foreground information is employed as the main information while the background information is treated as a secondary one. Differently from~\cite{Yang2014}, we propose to extract the foreground mask with a more powerful segmentation model based on Deep Decompositional Network (DDN)~\cite{Luo:ICCV:2013}. 

\begin{figure}[thpb]
	\centering
	\subfigure{\includegraphics[height=1.5cm]{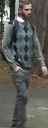}	}\hspace{-0.2cm}
	\subfigure{\includegraphics[height=1.5cm]{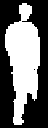}	}
	\subfigure{\includegraphics[height=1.5cm]{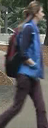}	}\hspace{-0.2cm}
	\subfigure{\includegraphics[height=1.5cm]{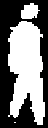}	}
	\subfigure{\includegraphics[height=1.5cm]{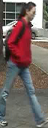}	}\hspace{-0.2cm}
	\subfigure{\includegraphics[height=1.5cm]{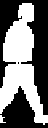}	}
	\subfigure{\includegraphics[height=1.5cm]{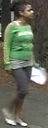}	}\hspace{-0.2cm}
	\subfigure{\includegraphics[height=1.5cm]{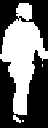}	}
	\subfigure{\includegraphics[height=1.5cm]{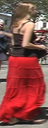}	}\hspace{-0.2cm}
	\subfigure{\includegraphics[height=1.5cm]{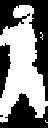}	}
	\caption{Input images and respective binary masks obtained using~\cite{Luo:ICCV:2013}.}
	\label{ddnimgs}
\end{figure}

The DDN was developed to tackle the problem of pedestrian parsing, and designed to segment pedestrian images into semantic regions, such as hair, head, body, arms, and legs. It directly maps low-level visual features (HOG) to the label maps of body parts, being able to accurately estimate complex pose variations while being robust to occlusions and background clutters. In a nutshell, DDN jointly estimates occluded regions and segments body parts by stacking three types of hidden layers: occlusion estimation layers, completion layers, and decomposition layers. The occlusion estimation layers estimate a binary mask, indicating which part of a pedestrian is invisible. The completion layers synthesize low-level features of the invisible part from the original features and the occlusion mask. The decomposition layers directly transform the synthesized visual features to label maps. Fig.~\ref{ddnimgs} illustrates some binary masks automatically obtained using~\cite{Luo:ICCV:2013}\footnote{Code available at~\url{http://mmlab.ie.cuhk.edu.hk/projects/luoWTiccv2013DDN/}}.

\subsubsection{Gaussian Of Gaussian (GOG)}\label{sec:gog}

Matsukawa et al.~\cite{Matsukawa:cvpr16} proposed a region descriptor based on hierarchical Gaussian distribution of pixel feature. In their work, local patches inside a region are densely extracted and the region is regarded as a set of local patches. The region is modeled as a set of multiple Gaussian distributions, each of them representing the appearance of one local patch. The characteristics of the set of patch Gaussians are again described by another Gaussian distribution (defined as \textit{a region Gaussian}). The parameters of the region Gaussian are then used as feature vector to represent the region. The GOG descriptor provides a consistent way to generate discriminative and robust features that describe color and textural (\textit{e.g.}, gradient magnitudes along different directions) information simultaneously. 

In our work, we adopted the GOG$_{F}$ (Fusion) descriptor\footnote{Available at \url{http://www.i.kyushu-u.ac.jp/~matsukawa/ReID.html}}, extracted as described in~\cite{Matsukawa:cvpr16}, which concatenates different GOG descriptors generated from different colorspaces (RGB, LAB, HSV and normalized \textit{rgb}). Thus, a new visual cue is proposed, $\mathbf{C}_7$. As before, obtained descriptor is normalized (to have unit $L_2$ norm). The resulting descriptor is then reduced by means of PCA before final normalization. Note that, as the final representation is a concatenation of local features, it will be used in Eq.~\ref{finalsimilarity} as a global descriptor (similarly as $\mathbf{C}_8$, described next).

\subsubsection{Deep feature~\cite{Wu:WACV:2016}} Feature Fusion Net (FFN) is used to allow deep feature representation in the adopted framework, as it demonstrated to be very effective in person re-identification tasks. FFN consists of two parts. The first part deals with traditional convolution, pooling and activation neurons for input images. The second part of the network processes additional hand-crafted feature representations of the same image. Both, CNN features and the hand-crafted features are followed by a fully connected layer and then linked together in order to produce a full-fledge image description from the last convolutional layer.

Regarding the hand-crafted features, authors first modified the Ensemble of Local Features (ELF)~\cite{Gray:ECCV:2008} by improving the color space and stripe division (denoted as ELF16). Input images are equally partitioned into 16 horizontal stripes, and the features are composed of color features including RGB, HSV, LAB, XYZ, YCbCr and NTSC, and texture features including Gabor, Schimid and LBP. A 16D histogram is extracted for each channel and then normalized by $L_1$ norm. All histograms are concatenated together to form a single vector. 
The FFN was then trained on the Market-1501~\cite{Zheng:ICCV:2015} dataset, which is the largest public person re-identification dataset up to date, composed of $38195$ images from $1501$ identities.

The authors of~\cite{Wu:WACV:2016} also mention that even though the proposed CNN-based feature performs better when compared to LOMO~\cite{Liao:CVPR:2015} features, the combination of both kind of features demonstrates to have higher discriminative power. Thus, the concatenation of both (CNN-based feat.+LOMO) is defined in their work as the final representation (denoted in our work, from now, by just \textit{Deep feature}. 
We also apply PCA (as previously mentioned), to reduce the dimensionality of the resulting Deep feature, which is then normalized by $L_2$ norm. This final representation is used as another complementary cue ($\mathbf{C}_8$). Note that $\mathbf{C}_8$ composes a representation for the whole image, so it will be only used as a global descriptor.

\subsubsection{Integrating complementary features}\label{featintegration}

To integrate the new and complementary features, we compute different similarity measures using different feature representations (for each pair of images being compared), which are then exploited next by the post-ranking and ranking aggregation strategies. 
To be specific, we compute $s^{final}_i(\mathbf{x}_a^{F_i},\mathbf{x}_b^{F_i})$, defined in Eq.~\ref{finalsimilarity}, where $F_i$ are different feature representations, described in Table~\ref{table:featrepresentations}. We refer \{$F_7,...,F_{12}$\} as simplified versions of \{$F_1,...,F_6$\}.

\begin{table}[htbp]
	\caption{Summary of the adopted feature representations $F_i$ ($F_0$ is the baseline~\cite{Chen:CVPR:2016}). G, L and GL indicate, respectively, if the visual cue is applied just on the global, local or both parts of the Eq.~\ref{finalsimilarity}.}
	\label{table:featrepresentations}
	\footnotesize
	\begin{center}
		\begin{tabular}{|l|c|c|c|c|c|c|c|c|}
			\hline
            & \multicolumn{4} {|c|}  {\textit{baseline cues}} & \multicolumn{2} {|c|}{\textit{scncd}} &{\textit{gog}}  & {\textit{deep}}\\
    \cline{2-9}
			& $\mathbf{C}_1$ & $\mathbf{C}_2$ & $\mathbf{C}_3$ & $\mathbf{C}_4$ & $\mathbf{C}_5$ & $\mathbf{C}_6$ & $\mathbf{C}_7$ & $\mathbf{C}_8$\\
			\hline
			$F_0$ & GL & GL & GL & GL & - & - & -  & -\\
			\hline
			$F_1$ & GL & GL & GL & GL & - & - & G & -\\
			\hline
			$F_2$ & GL & GL & GL & GL & - & - & -  & G\\
			\hline
			$F_3$ & GL & GL & GL & GL & - & - & G & G\\
			\hline			
			$F_4$ & - & - & - & - & GL & GL & G & -\\
			\hline			
			$F_5$ & - & - & - & - & GL & GL & - & G\\
			\hline			
			$F_6$ & - & - & - & - & GL & GL & G  & G\\
			\hline			
			$F_7$ & L & L & L & L & - & - & G &  -\\
			\hline
			$F_8$ & L & L & L & L & - & - & - &  G\\
			\hline
			$F_9$ & L & L & L & L & - & - & G &  G\\
			\hline			
			$F_{10}$ & - & - & - & - & L & L & G  & -\\
			\hline			
			$F_{11}$ & - & - & - & - & L & L & -  & G\\
			\hline			
			$F_{12}$ & - & - & - & - & L & L & G  & G\\
			\hline			
		\end{tabular}
	\end{center}
\end{table}

\subsection{Post-ranking based on DCIA}\label{post-ranking}

According to Garcia et al.\cite{Garcia:ICCV:2015,Garcia:2017}, additional ranking inspections on the ranked matches can be applied to refine the output in such a way that the correct match will have higher probability to be found in the first ranks. To this end, they proposed the Discriminant context information Analysis (DCIA) method, which is built under the definition of content and context information. The content information is the set of gallery images that have low dissimilarity with respect to the probe. The context information is the set of gallery images that have low dissimilarity with both the probe and an image of the content information. In this subsection we introduce basic concepts related to their method as well as describe how post-ranking is performed.

\subsubsection{Definitions}

Let $\mathcal{A}=\{\mathbf{I}_p^A\}_{p=1}^N$ be the set of $N$ probe images and $\mathcal{B}=\{\mathbf{I}_g^B\}_{g=1}^M$ be the set of $M$ gallery images. Given a probe image $\mathbf{I}_p^A$, its initial ranking is defined as $\mathcal{R}_p=\{\mathbf{I}_i^B\}_{i=1}^M$, where the gallery images $\mathbf{I}_i^B$ are sorted depending on the dissimilarity to the probe. In other words, $d(\mathbf{I}_p^A,\mathbf{I}_i^B) < d(\mathbf{I}_p^A,\mathbf{I}_{i+1}^B)$, where $d(\cdot,\cdot)$ is a suitable dissimilarity measure (\textit{i.e.}, as defined in Eq.~\ref{finalsimilarity}) and $i$ goes from 1 to $M-1$.  $\mathbcal{R}=\{\mathcal{R}_p^N\}_{p=1}^N$ denotes the set of such initial rankings computed for the $N$ probes.

\subsubsection{Content Information}

The content information is defined as the set of features extracted from the correlated matches, \textit{i.e.}, a subset of gallery images $\mathcal{B}^{cn} \subseteq \mathcal{B}$ present in the fist ranks and which are likely to share visual ambiguities with the probe. Elements in such a set are selected from the top $m$ positions in the initial ranking $\mathcal{R}_p$ which have matching distance less than a specific threshold $Th$. The $m$ value, as well as the adopted threshold, are dynamically computed for each probe image, based on the shape of dissimilarities vs rank plots (see~\cite{Garcia:ICCV:2015} for additional details). Thus, the set of $m$ correlated matches equals $\mathcal{B}^{cn}=\{\mathbf{I}_i^B|d(\mathbf{I}_p^A,\mathbf{I}_i^B)\leq Th\}$. Therefore, the content set $\mathcal{C}_p^{cn}=\{\mathbf{x}_1^{cn},...,\mathbf{x}_m^{cn}\}$ contains the $m$ feature vectors extracted from the correlated matches in $\mathcal{B}^{cn}$. Notice that, only images in $\mathcal{C}_p^{cn}$ are re-ranked.

\subsubsection{Context Information}

The context information is given by the $K$-common nearest neighbors of the probe and a correlated match. Given $\mathbf{I}_p^A$, its respective context set is extracted by exploiting $\mathcal{C}_p^{cn}$. First, the initial rank list $\mathcal{R}_g$ is computed for each correlated matching image $\mathbf{I}_g^B \in \mathcal{C}_p^{cn}$ by evaluating its similarity with images in the gallery set $\mathcal{B}^*=(\mathcal{B} \backslash \mathbf{I}_g^B)$ using model parameters ($\phi$ and  $\mathbf{W}$) and distance $d(\cdot,\cdot)$, \textit{i.e.}, as defined in Eq.~\ref{finalsimilarity}. Then, given $\mathcal{R}_g$, we compute the top $m_g$ positions which have matching distance less than $Th_g$ (being $m_g$ and $Th_g$ computed in the same way as $m$ and $Th$, respectively). These elements represent the images that have high similarity with both the probe $\mathbf{I}_p^A$ and the correlated match $\mathbf{I}_g^B$. The context information is extracted from the $K$-common context matches. Feature vectors extracted from such images form the context information set $\mathcal{C}_p^{cx} =\{\mathbf{x}_1^{cx},...,\mathbf{x}_n^{cx}\}$, where $n=K$. Finally, $\mathcal{C}_p^{cx}$ is updated by removing images that are in duplicate with $\mathcal{C}_p^{cn}$. The hard threshold $K$ was set experimentally to $K=13$ as in~\cite{Garcia:ICCV:2015}. 

Nevertheless, we observed the respective $K$-common context matches are obtained, for some probe images, from a flat histogram. In this case, the $K$-common context matches might be imprecisely obtained, mainly when the number of images that compose the histogram are greater than $K$. Thus, we introduced a new condition to also consider the similarity from the correlated context match to the probe image in order get the $K$-common context matches most similar to $\mathbf{I}_p^A$.

\subsubsection{Discriminative Information Analysis}

Given a probe image $\mathbf{I}_p^A$, let $\mathcal{D}_p =\{\mathbf{x}_p,\mathcal{C}_p^{cn},\mathcal{C}_p^{cx}\}$ be the set composed of its feature vector and of feature vectors obtained in the content and context information. $\mathcal{D}_p$ is redefined as a feature matrix $\mathbf{D}_p \in \mathbb{R}^{d\times l}$ with zero mean, where $l=1+m+n$ is the number of vectors. Let $\mathbf{P}\in \mathbb{R}^{d\times k}$ be the first $k$ components of $\mathbf{D}_p$ selected to represent the common appearance subspace. Thus, the discriminant information can be obtained as

\begin{equation}
\mathbf{D}_p^* = \mathbf{D}_p - \mathbf{P}\mathbf{P}^T\mathbf{D}_p,
\label{pca:dcia}
\end{equation}
where each column of $\mathbf{D}_p^*$ represents a discriminant feature vector $\mathbf{x}^*$. Differently from~\cite{Garcia:ICCV:2015}, where $k$ principal components corresponding to the 55\% ($k=0.55$) of energy of the set of feature vectors have been used to represent the common appearance subspace, in this work we empirically defined $k=0.35$.

\subsubsection{Re-ranking Training}

The DCIA is first applied to the train set $\mathcal{I}_{Tr}$. More specifically, it is applied to each ranking $\mathcal{R}_p^{Tr} \in \mathbcal{R}^{Tr}$. As result, the discriminant feature vectors $\mathbf{x}_p^{*A}$ and $\mathbf{x}_g^{*B} \in \mathbf{D}_p^{*Tr}$ are obtained for each probe image $p$. The resulting sets $\mathbf{x}_{Tr}^{*A}$ and $\mathbf{x}_{Tr}^{*B}$ together with the pairwise labels are used to learn the new model parameters $\phi^*$ and  $\mathbf{W}^*$.

\subsubsection{Post-ranking Optimization}

Given a test rank in $\mathbcal{R}$, the DCIA is performed to obtain the discriminative test feature vectors $\mathbf{x}_p^{*A}$ and $\mathbf{x}_g^{*B}$. Then, the set of such vectors \{$\mathbf{x}^{*A},\mathbf{x}^{*B}$\} is evaluated by the new model parameters $\phi^*$ and $\mathbf{W}^*$. The obtained distances are used to re-rank the correlated matches, hence to compute the final ranking $\mathbf{\tilde{\mathbcal{R}}}$. Such procedure is performed for each feature representation.

\subsection{Ranking Aggregation Strategy}\label{rankagg}

We propose to explore different feature representations to obtain complementary ranking lists and combine them using the Stuart ranking aggregation method~\cite{Stuart:Science:2003}.
The Stuart ranking aggregation method, which was originally designed to define a gene-coexpression network over DNA microarrays from humans, flies, worms, and yeast, is a probabilistic method based on order statistics to evaluate the probability of observing a particular configuration of ranks across the different organisms, even when there are irrelevant and noise inputs. The significance of the interactions in the network is verified by means of a variety of statistical tests~\footnote{An optimized solution of~\cite{Stuart:Science:2003} is presented in~\cite{Kodle:BIOINF:2012}.}.

Let first denote by $\oplus$ the aggregation operator, for instance if $\mathbf{\tilde{\mathbcal{R}}}_n = \mathbf{\tilde{\mathbcal{R}}}_1 \oplus \mathbf{\tilde{\mathbcal{R}}}_2 \oplus ... \oplus \mathbf{\tilde{\mathbcal{R}}}_{n-1}$, then $\mathbf{\tilde{\mathbcal{R}}}_n$ is a ranking list computed by the aggregation of ranking lists from $\mathbf{\tilde{\mathbcal{R}}}_1$ to $\mathbf{\tilde{\mathbcal{R}}}_{n-1}$. As we use different descriptors to represent each image, and have adopted a strategy in which we can measure the similarity $s^{final}_i(\mathbf{x}_a,\mathbf{x}_b)$ of image pairs using different ways (Sec.~\ref{featintegration}), we are also able to compute different ranking lists for each probe image and gallery set
, as illustrated in Fig.~\ref{overview}. Moreover, a \textit{tunning} strategy is performed on the final aggregation in order to improve the results, as explained next.

Consider the list of complementary features $\mathcal{L} = $\{$F_1,...,F_{12}$\}, sorted based on average top-1 rank recognition rate (obtained from a validation set), where $\mathcal{L}_i$ has better accuracy than $\mathcal{L}_{i+1}$. Thus, we can aggregate the ranking lists that obtained higher accuracies (\textit{i.e.}, the best-$n$ feature representations) and ignore ``poor'' ranking lists that may push down the final result/aggregation. Such procedure is performed by the aggregation of the ranking lists of the best-$2$ feature representations (\textit{i.e.}, \{$\mathcal{L}_1$,$\mathcal{L}_2$\}), best-$3$ (\{$\mathcal{L}_1,...,\mathcal{L}_3$\}), and so on, up to the aggregation of the whole list (\textit{i.e.}, best-$12$, \{$\mathcal{L}_1,...,\mathcal{L}_{12}$\}).

\section{EXPERIMENTAL RESULTS} \label{experimentalresults}

In order to demonstrate the effectiveness of the proposed model, this section presents experimental results on three broadly employed public datasets for person re-identification, \textit{i.e.}, VIPeR~\cite{Gray:PETS:2007}, PRID450s~\cite{Roth:ACVPR:2014} and CUHK01~\cite{Li:ACCV:2012}. Five case studies were performed. First, (i) the proposed model was compared against state-of-the-art person re-identification models using a well known evaluation protocol (Sec.~\ref{sotacompsec}). Then, we decomposed the proposed complementary features and performed the following experiments: (ii) influence of the background/foreground information within SCNCD (Sec.~\ref{exp:bkg}); (iii) accuracy performance obtained by each complementary feature (Sec.~\ref{complementaryfeat}); (iv) improvements obtained by post-ranking (Sec.~\ref{post-ranking-analysis}). Finally, (v) the best-$n$ \textit{tunning} strategy for rank-aggregation was analyzed (Sec.~\ref{rank-agg-analysis}).

\begin{figure}[thpb]
	\centering
	\subfigure[VIPeR]{\includegraphics[height=3.5cm]{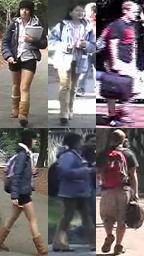}	}
	\subfigure[PRID450s]{\includegraphics[height=3.5cm]{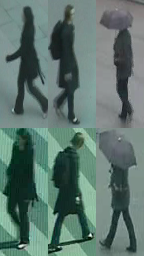}	}
	\subfigure[CUHK01]{\includegraphics[height=3.5cm]{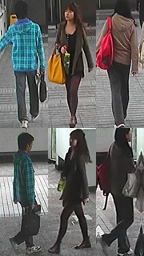}	}
	\caption{Sample images of the adopted datasets. Images on the same column represent the same person.}
	\label{fig:dataset}
\end{figure}

The adopted datasets are presented in two disjoint camera views, with significant misalignment, light changes and body part distortion. Table~\ref{table:datasets} summarizes the three datasets. Challenging image samples (due to illumination problems, pose variation, occlusions or even by high similarity between different people) are illustrated in Fig.~\ref{fig:dataset}.

\begin{table}[htbp]
	\caption{Summary of the adopted datasets.}
	\label{table:datasets}
	\footnotesize
	\begin{center}
		\begin{tabular}{|c|c|c|c|}
			\hline
			& VIPeR & PRID450s & CUHK01 \\
			\hline
			Images & 1264 & 900 & 3884\\
			\hline
			Individuals (ID) & 632 & 450 & 971\\
			\hline
			Images per ID (per view) & 1 & 1 & 2\\
			\hline
		\end{tabular}
	\end{center}
\end{table}

\subsection{Evaluation Protocol}\label{evalprotocol}
Our experiments follow the evaluation protocol defined in~\cite{Paisitkriangkrai:CVPR:2015} for a single-shot scenario, \textit{i.e.} we randomly partitioned each dataset into two parts, 50\% for training and 50\% for testing, without overlap on person identities. As the CUHK01 dataset contains $971$ individuals, $485$ of them were randomly sampled for training and the rest for testing, as in~\cite{Zhang:CVPR:2016}. Images from camera A are used as probe and those from camera B as gallery. For the CUHK01 dataset, in which each individual has two images per camera view, we randomly selected one image of the individual taken from the camera A as the probe image and one image of the same individual taken from the camera B as the gallery image. For all evaluated datasets, each probe image is matched with every image in gallery and the rank of correct match is obtained. This procedure is repeated $10$ times and the average of Cumulative Matching Characteristic (CMC) curves across $10$ partitions is reported.

\subsection{Case 1: State-of-the-art comparison}\label{sotacompsec}

This experiment compares the overall accuracy performance of the proposed model in relation to the state-of-the-art. Different feature representations were integrated, as described in Sec.~\ref{featintegration}, followed by the post-ranking approach (described in Sec.\ref{post-ranking}) and ranking aggregation strategies (described in Sec.~\ref{rankagg}). Table~\ref{sotacomp} summarizes the obtained results. 
	
\begin{table}[htbp]
	\caption{State-of-the-art comparison. Top Matching Rank (\%) on the three adopted datasets.}
	\label{sotacomp}
	\footnotesize
	\begin{center}
		\begin{tabular}{|c|c|c|c|c|}
			\hline
			Rank & 1 & 5 & 10 & 20\\
			\hline
			\hline
			\multicolumn{5}{|c|}{\textbf{VIPeR}}\\
			\hline
			\hline
			\textbf{Our best-$10$} & \textbf{67.21} & \textbf{87.78} & 93.39 & \textbf{97.82}\\
			\hline
			\textbf{Our best-$12$} & 66.83 & 87.78 & \textbf{93.41} & 97.72\\
			\hline
			DCIA~\cite{Garcia:2017} & 64.78 & 76.85 & 86.88 & 94.77\\
			\hline
			Re-ranking~\cite{Mirmahboub:2017} & 59.46 & 86.68 & 93.39 & 97.63\\
			\hline
			SSM~\cite{Bai:CVPR17} & 53.73 & - & 91.49 &  96.08\\
			\hline
			SCSP~\cite{Chen:CVPR:2016} & 53.54 & 82.59 & 91.49 & 96.65\\
			\hline
			Deep+LOMO~\cite{Wu:WACV:2016} & 51.06 & 81.01 & 91.39 & 96.90\\
			\hline
			CAN-VGG16~\cite{Liu:TIP17} & 47.20 & 79.20 & 89.20 &  95.80\\
			\hline
			Mirror~\cite{Chen:IJCAI:2015} & 42.97 & 75.82 & 87.28 & 94.84\\
			\hline
			LSSCDL~\cite{Zhang:CVPR:2016} & 42.66 & - & 84.27 & 91.93\\
			\hline
			\hline
			\multicolumn{5}{|c|}{\textbf{PRID450s}}\\
			\hline
			\hline
			\textbf{Our best-$3$} & \textbf{75.64} & \textbf{93.38} & 96.44 & 98.22\\
			\hline
			\textbf{Our best-$12$} & 73.91 & 92.58 & 95.87 & 97.87\\
			\hline
			SSM~\cite{Bai:CVPR17} & 72.98 & - & \textbf{96.76} &  \textbf{99.11}\\
			\hline
			Deep+LOMO~\cite{Wu:WACV:2016} & 66.62 & 86.84 & 92.84 &  96.89\\
			\hline
			LSSCDL~\cite{Zhang:CVPR:2016} & 60.49 & - & 88.58 & 93.60\\
			\hline
			Mirror~\cite{Chen:IJCAI:2015} & 55.42 & 79.29 & 87.82 &  93.87\\
			\hline
			\hline
			\multicolumn{5}{|c|}{\textbf{CUHK01}}\\
			\hline
			\hline
			CAN-VGG16~\cite{Liu:TIP17} & \textbf{67.20} & 87.30 & \textbf{92.50} &  \textbf{97.20}\\
			\hline
			\textbf{Our best-$3$} & 66.91 & 86.95 & 92.12 & 95.7 \\
			\hline
			LSSCDL~\cite{Zhang:CVPR:2016} & 65.97 & $\approx$ \textbf{88.0} & $\approx$ 92.0 &  $\approx$ 96.0\\
			\hline
			\textbf{Our best-$12$} & 64.28 & 85.21 & 90.78 & 95.00 \\
			\hline
			Deep+LOMO~\cite{Wu:WACV:2016} & 55.51 & 78.40 & 83.68 &  92.59\\
			\hline
			Mirror~\cite{Chen:IJCAI:2015} & 40.40 & 64.63 & 75.34 &  84.08\\
			\hline
		\end{tabular}
	\end{center}
\end{table}

As it can be seen in Table~\ref{sotacomp}, the proposed model outperforms the state-of-the-art on both VIPeR and PRID450s datasets, and achieved competitive results on CUHK01 dataset. Some other works obtained better results than ours on CUHK01 dataset, however, they were not included in this comparison as they either use a different evaluation protocol~\cite{Chen:TPAMI17} or include additional data in the train set~\cite{Zhao:CVPR17} (\textit{i.e.}, CUHK03 database, which was captured in the same environment as CUHK01 and could benefit when CUHK01 is evaluated as both share similar features).

We can also observe the CAN-VGG16 method~\cite{Liu:TIP17}, which obtained promising results on CUHK01 dataset, were outperformed by the proposed model on VIPeR dataset by a significant margin. The slow performance related to CAN-VGG16 method on VIPeR database is because the size of training set of VIPeR is so small.

\subsection{Case 2: Background information within SCNCD}\label{exp:bkg}

\begin{figure}[htbp]
	\centering
	\includegraphics[height=4.2cm]{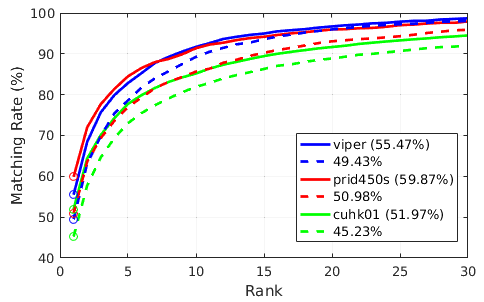}
	\caption{Accuracy performances based on SCNCD (\textit{i.e.}, using only $\mathbf{C}_5$ and $\mathbf{C}_6$), with and without background/foreground information (solid and dashed lines, respectively). Top-1 rank values for each case are also provided.}
	\label{cmcbkg}
\end{figure}

This experiment analyzes the accuracy performance of the background/foreground information within SCNCD (Sec.~\ref{contextinformation}), before the employment of the post-ranking approach (Sec.~\ref{post-ranking}). To this end, we set up the adopted framework to load only the following visual cues, $\mathbf{C}_5$ and $\mathbf{C}_6$ (detailed in Sec.~\ref{scncdfeat}, \textit{i.e.}, without deep features), both without and with background/foreground information. Fig.~\ref{cmcbkg} shows the CMC curves obtained for this experiment (for the first rank values). As it can be observed, the background/foreground information significantly improved the overall accuracy on the three evaluated datasets, being effective to remove the background noise. Yang et al.~\cite{Yang2014} obtained same conclusion when evaluating both representations (image-foreground and image-only) on VIPeR and PRID450s datasets. However, differently from their work, in which the evaluation was performed using only RGB information combined with the segmentation model proposed in~\cite{Jojic:CVPR:2009} and the KISSME~\cite{KISSME:CVPR:2012} metric learning, we adopted a more powerful segmentation strategy, as well as a different similarity function. 

It can also be noticed from Fig.~\ref{cmcbkg} that the proposed feature representation based on SCNCD outperformed obtained results (for the VIPeR dataset) reported in~\cite{Chen:CVPR:2016} (see Table~\ref{sotacomp}).

\subsection{Case 3: Complementary feature representations}\label{complementaryfeat}	

This experiment evaluated the complementary features individually (before post-ranking). Each proposed feature representation was integrated as detailed in Sec.~\ref{featintegration}, and $s^{final}_i$ is adopted as the similarity function related to each representation $F_i$. Obtained results are shown in Fig.~\ref{featrep} in terms of top-1 rank recognition rate, from where we can make the following observations:	

\begin{figure*}[thpb]
	\centering
	\includegraphics[width=0.95\textwidth]{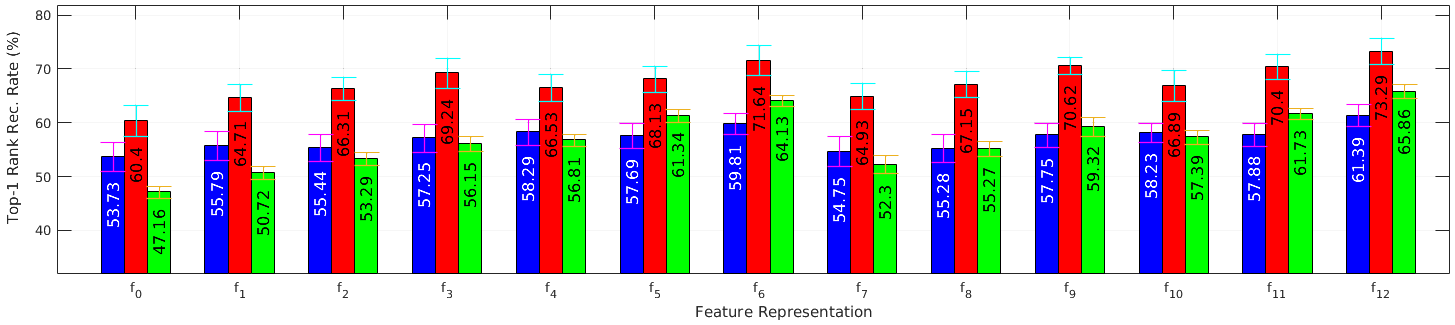}
	\caption{Accuracy performance obtained for each feature representation $F_i$ (before post-ranking), for the VIPeR (blue), PRID450s (red) and CUHK01 (green) datasets.}
	\label{featrep}
\end{figure*}

\begin{itemize}
	\item All complementary features outperformed the baseline feature representation $F_0$.
	\item The benefits of including GOG feature into the baseline feature representation can be observed if we compare overall results obtained for the respective pair of feature representations $\langle F_0, F_1\rangle$. Similarly, the benefits of including deep feature can be observed if we compare results for the pair $\langle F_0, F_2\rangle$. 
	\item The inclusion of both visual cues based on GOG and deep features, either on the baseline feature representation or on the proposed complementary features, can be highlighted if we compare overall obtained results for the respective pairs $\langle (F_1, F_2), F_3\rangle$ and $\langle (F_4, F_5), F_6\rangle$, as well as in relation to their respective simplified versions $\langle (F_7, F_8), F_9\rangle$ and $\langle (F_{10}, F_{11}), F_{12}\rangle$.
	\item The simplified versions of the complementary features $F_9$ and $F_{12}$ obtained better accuracy than their respective complete representations ($F_3$ to $F_6$), indicating that the proposed simplification still has strong discriminative power for person re-identification applications, while requiring less computation resources.
	\item $F_6$ and $F_{12}$, which exploits SCNCD (with background/foreground information), GOG and deep features, obtained the best overall accuracy performance in the three adopted datasets.
\end{itemize}

The previously mentioned observations indicate that the proposed complementary feature representations have strong discriminative power in person re-identification applications, mainly when combined through ranking aggregation, as shown in Sec.~\ref{sotacompsec}. As it will be described in Sec.~\ref{post-ranking-analysis}, accuracies obtained by such complementary features can also be improved by the post-ranking approach. In addition, different integration strategies (from those described in Sec.~\ref{featintegration}) were also evaluated in other experiments (\textit{e.g.}, the integration of all features, $\mathbf{C}_1$ to $\mathbf{C}_9$, using the simplified and complete representations), however, no significant accuracy performance improvements were observed.

\subsection{Case 4: Post-ranking analysis}\label{post-ranking-analysis}

\begin{figure}[htbp]
	\centering
	\includegraphics[height=3.5cm]{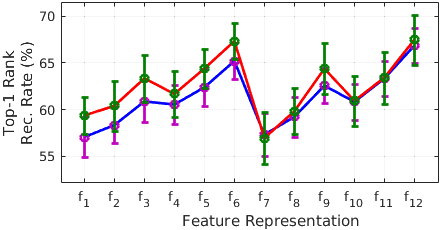}
	\caption{Average accuracy performance obtained for each feature representation, with (red) and without (blue) post-ranking.}
	\label{fig:post-avg-feat}
\end{figure}

In this section we analyze the results obtained by the DCIA method applied to the proposed framework. Our experiments were performed without visual expansion (employed in~\cite{Garcia:2017}), which synthesizes the probe into the gallery feature space aiming to reduce feature inconsistency. We avoided using this procedure because of its high computational requirements and, as commented by the authors, obtained accuracies were improved by less than 1\%. Figure~\ref{fig:post-avg-feat} shows results (averaged per database) for each complementary feature, before and after post-ranking. As we can observe, the post-raking approach improved overall results.

\begin{figure}[thpb]
	\centering
	\includegraphics[height=2.5cm]{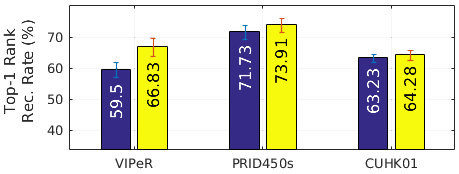}
	\caption{Average accuracy performance (after rank-aggregation) without/with post-ranking, respectively. Values obtained using the best-$12$ aggregation strategy described in Sec.~\ref{rank-agg-analysis}.}
	\label{fig:post-rank}
\end{figure}

As related in~\cite{Garcia:2017}, one limitation of the DCIA method is that sometimes the true match is not included in the content set. In this case, it will not be re-ranked. In addition, some images might move to higher rank positions after post-ranking. However, in general, it improved more than deteriorated final results. Fig.~\ref{fig:post-rank} shows the obtained improvement by the post-ranking approach (after rank-aggregation) on the three employed databases.

Table~\ref{table:post-rank} shows statistics related to the post-ranking approach (before ranking aggregation), considering 10 different runs and all complementary features $\{F_1,...,F_{12}\}$. First, values were averaged in relation to each feature representation and number of runs. Then, the average value per database was computed.

\begin{table}[htbp]
	\caption{Post-ranking statistics: mean average and standard deviation (\%) of: (i) probe images included in the content set, (ii) improved results, (iii) from the improved results, the ones that were moved to top-1 rank position, (iv) unchanged ranks and (v) images that were moved to higher rank positions (worsen).}
	\label{table:post-rank}
	\footnotesize
	\begin{center}
		\begin{tabular}{|c|c|c|c|}
			\hline
			& VIPeR & PRID450s & CUHK01 \\
			\hline
			i & 76.3 $\pm$2.8 & 81.2 $\pm$1.7 & 71.3 $\pm$1.4\\
			\hline
			ii & 11.4 $\pm$2.4 & ~7.0 $\pm$1.8 & ~6.4 $\pm$1.1\\
            \hline
			iii & 78.1 $\pm$6.8 & 84.0 $\pm$9.1 & 81.3 $\pm$7.3\\
            \hline
			iv & 80.5 $\pm$2.9 & 86.3 $\pm$2.5 & 85.7 $\pm$1.8\\
            \hline
			v & ~8.1 $\pm$2.1 & ~6.6 $\pm$1.8 & ~7.9 $\pm$1.5\\
			\hline
		\end{tabular}
	\end{center}
\end{table}

From Table~\ref{table:post-rank}, we can observe that a high percentage of images which rank position was improved (ii), were moved to top-1 rank position (iii). On the other hand, few images were undesired moved to higher rank positions (v). However, post-ranking improved overall results.

\subsection{Case 5: best-$n$ tunning rank-aggregation strategy}\label{rank-agg-analysis}

\begin{figure}[htbp]
	\centering
	\includegraphics[height=3.2cm]{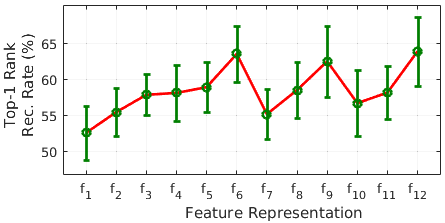}
	\caption{Average accuracy performance obtained for each feature representation on the validation set.}
	\label{fig:post-avg-feat-validation}
\end{figure}

In this experiment, the train set of each dataset was divided into 50\% train and 50\% validation. Fig.~\ref{fig:post-avg-feat-validation} shows obtained results on the validation set (averaged per database) for each complementary feature (with post-ranking). Table~\ref{table:rank-agg} shows obtained results, on the test set, from the aggregation of the best-$n$ feature representations, for $n=2$ to $12$. As we can observe from Table~\ref{table:rank-agg}, the inclusion of additional feature representations do not always increase accuracy performance, as well as there is not an overall feature representation that fits all databases.

\begin{table}[htbp]
	\caption{Top-1 rank recognition rate (\%) after ranking-aggregation, using the ``best-$n$'' feature representations.}
	\label{table:rank-agg}
	\footnotesize
	\begin{center}
		\begin{tabular}{|c|c|c|c|}
			\hline
			best-$n$ & VIPeR & PRID450s & CUHK01\\
			\hline
			2 & 64.75 &  73.87 &  64.90\\
			\hline
			3 & 66.99 &  \textbf{75.64} &  \textbf{66.91}\\
			\hline
			4 & 66.87 &  75.24 &  66.75\\
			\hline
			5 & 66.83 &  74.56 &  66.30\\
			\hline
			6 & 66.77 &  74.31 &  66.69\\
			\hline
			7 & 66.33 &  74.18 &  66.34\\
			\hline
			8 & 66.83 &  74.98 &  65.05\\
			\hline
			9 & 67.15 &  75.07  & 66.13\\
			\hline
			10 & \textbf{67.21} &  74.40 &  65.49\\
			\hline
			11 & 66.77 &  73.95 &  65.10\\
			\hline
			12 & 66.83 &  73.91 &  64.28\\
			\hline
		\end{tabular}
	\end{center}
\end{table}

\subsection{Computational cost}
We adapted the MATLAB implementation provided in~\cite{Chen:CVPR:2016} to consider the proposed complementary features. Computational costs shown in Table~\ref{table:comp-cost} were obtained using the VIPeR dataset. The complete representations \{$F_3,F_6$\}, which explore different visual cues, as well as their respective simplified versions \{$F_9,F_{12}$\} were analyzed\footnote{Using a 2.30GHz Intel Core i7 CPU and 8Gb of memory, without considering I/O procedures and image resize operations.}.

\begin{table}[htbp]
	\caption{Average computational cost obtained from 10 runs, to process (train and test) the whole VIPeR database (316 train and 316 test images), taking into account different features representations.}
	\label{table:comp-cost}
	\footnotesize
	\begin{center}
		\begin{tabular}{|c|c|c|c|}
			\hline
			    & Total & Post-rank & Test\\
			    \hline
			$F_3$ & 16.9m $\pm$4.2 & 6.1m $\pm$0.4 & 4.17s $\pm$1.5\\
			\hline
			$F_6$ & ~5.0m $\pm$0.7 & 3.3m $\pm$0.3 & 2.19s $\pm$0.6\\
			\hline
			$F_9$ & ~7.3m $\pm$1.9 & 4.1m $\pm$0.3 & 2.93s $\pm$1.1\\
			\hline
			$F_{12}$ & ~3.7m $\pm$0.4 & 2.5m $\pm$0.4 & 1.91s $\pm$0.5\\
			\hline
		\end{tabular}
	\end{center}
\end{table}

Average computational time to run the ranking-aggregation, using the best-$7$ strategy (described in Sec.~\ref{rank-agg-analysis}) was 3.9m $\pm$0.4.

\section{Conclusion}\label{finalconsiderations}

In this work we exploited different feature representations, combined with a post-ranking and ranking aggregation strategies, to advance the state-of-the-art in person re-identification. Our model was built on a framework combining similarity learning metric with spatial constraints. The proposed complementary features demonstrated to have strong discriminative power, as well as  to complement each other even when the simplified versions are employed. Different feature representations were analyzed individually and incrementally. The post-ranking approach demonstrated to be a powerful tool in person re-identification tasks, being able to improve initial results which could be further enhanced by the ranking-aggregation strategy. We show that handcrafted and deep features fusion enhance re-identification performance especially in domains where there is a reduced amount of available data. As a result, we improved the top-1 rank recognition by $2.43$\% and $2.66$\% on VIPeR and PRID450s datasets, respectively, as well as obtained competitive results on the CUHK01 database.

\section*{Acknowledgements}
This work has been partially supported by the Spanish projects TIN2015-66951-C2-2-R and TIN2016-74946-P (MINECO/FEDER, UE), by the European Comission Horizon 2020 granted project SEE.4C under call H2020-ICT-2015, by the CERCA Programme/Generalitat de Catalunya. We gratefully acknowledge the support of NVIDIA Corporation with the donation of the Titan Xp GPUs used for this research.

\bibliographystyle{elsarticle-num}

\begin{thebibliography}{10}
\expandafter\ifx\csname url\endcsname\relax
  \def\url#1{\texttt{#1}}\fi
\expandafter\ifx\csname urlprefix\endcsname\relax\def\urlprefix{URL }\fi
\expandafter\ifx\csname href\endcsname\relax
  \def\href#1#2{#2} \def\path#1{#1}\fi

\bibitem{Vezzani:2013}
R.~Vezzani, D.~Baltieri, R.~Cucchiara, People reidentification in surveillance
  and forensics: A survey, ACM Computing Surveys 46~(2) (2013) 29:1--29:37.

\bibitem{BedagkarGala:2014:IVC}
A.~Bedagkar-Gala, S.~K. Shah, A survey of approaches and trends in person
  re-identification, Image and Vision Computing 32~(4) (2014) 270 -- 286.

\bibitem{Survey:Springer:2014}
S.~Gong, M.~Cristani, S.~Yan, C.~L. (Eds.), Person Re-Identification, Advances
  in Computer Vision and Pattern Recognition. Springer, 2014.

\bibitem{Cheng:CVPR:2016}
D.~Cheng, Y.~Gong, S.~Zhou, J.~Wang, N.~Zheng, Person re-identification by
  multi-channel parts-based cnn with improved triplet loss function, in: CVPR,
  2016, pp. 1335--1344.

\bibitem{Chen:CVPR:2016}
D.~Chen, Z.~Yuan, B.~Chen, N.~Zheng, Similarity learning with spatial
  constraints for person re-identification, in: CVPR, 2016, pp. 1268--1277.

\bibitem{Yang2014}
Y.~Yang, J.~Yang, J.~Yan, S.~Liao, D.~Yi, S.~Z. Li, Salient color names for
  person re-identification, in: ECCV, 2014, pp. 536--551.

\bibitem{Chen:CVPR:2015}
D.~Chen, Z.~Yuan, G.~Hua, N.~Zheng, J.~Wang, Similarity learning on an explicit
  polynomial kernel feature map for person re-identification, in: CVPR, 2015,
  pp. 1565--1573.

\bibitem{Bazzani:CVIU:2013}
L.~Bazzani, M.~Cristani, V.~Murino, Symmetry-driven accumulation of local
  features for human characterization and re-identification, Computer Vision
  and Image Understanding 117~(2) (2013) 130--144.

\bibitem{KISSME:CVPR:2012}
M.~Köstinger, M.~Hirzer, P.~Wohlhart, P.~M. Roth, H.~Bischof, Large scale
  metric learning from equivalence constraints, in: CVPR, 2012, pp. 2288--2295.

\bibitem{Garcia:2017}
J.~Garc\'ia, N.~Martinel, A.~Gardel, I.~Bravo, G.~L. Foresti, C.~Micheloni,
  Discriminant context information analysis for post-ranking person
  re-identification, IEEE Transactions on Image Processing 26~(4) (2017)
  1650--1665.

\bibitem{Wu:WACV:2016}
S.~Wu, Y.~C. Chen, X.~Li, A.~C. Wu, J.~J. You, W.~S. Zheng, An enhanced deep
  feature representation for person re-identification, in: IEEE Winter Conf. on
  Applications of Computer Vision (WACV), 2016.

\bibitem{Chen:IJCAI:2015}
Y.-C. Chen, W.-S. Zheng, J.~Lai, Mirror representation for modeling
  view-specific transform in person re-identification, in: International Joint
  Conf. on Artificial Intelligence, 2015, pp. 3402--3408.

\bibitem{Xiao:CVPR:2016}
T.~Xiao, H.~Li, W.~Ouyang, X.~Wang, Learning deep feature representations with
  domain guided dropout for person re-identification, in: CVPR, 2016, pp.
  1249--1258.

\bibitem{Paisitkriangkrai:CVPR:2015}
S.~Paisitkriangkrai, C.~Shen, A.~van~den Hengel, Learning to rank in person
  re-identification with metric ensembles, in: CVPR, 2015, pp. 1846--1855.

\bibitem{Li:CVPR:2014}
W.~Li, R.~Zhao, T.~Xiao, X.~Wang, Deepreid: Deep filter pairing neural network
  for person re-identification, in: CVPR, 2014, pp. 152--159.

\bibitem{Gray:PETS:2007}
D.~Gray, S.~Brennan, H.~Tao, Evaluating appearance models for recognition,
  reacquisition, and tracking, in: IEEE Int. Workshop on Performance Evaluation
  for Tracking and Surveillance, 2007.

\bibitem{Prates:ICIP:2015}
R.~F. de~Carvalho~Prates, W.~R. Schwartz, {C}{B}{R}{A}: Color-based ranking
  aggregation for person re-identification, in: IEEE International Conference
  on Image Processing (ICIP), 2015, pp. 1975--1979.

\bibitem{Roth:ACVPR:2014}
P.~M. Roth, M.~Hirzer, M.~Koestinger, C.~Beleznai, H.~Bischof, Mahalanobis
  distance learning for person re-identification, in: S.~Gong, M.~Cristani,
  S.~Yan, C.~C. Loy (Eds.), Person Re-Identification, Springer, London, United
  Kingdom, 2014, pp. 247--267.

\bibitem{Li:ACCV:2012}
W.~Li, R.~Zhao, X.~Wang, Human reidentification with transferred metric
  learning, in: ACCV, 2012, pp. 31--44.

\bibitem{Dalal:CVPR:2005}
N.~Dalal, B.~Triggs, Histograms of oriented gradients for human detection, in:
  CVPR, 2005, pp. 886--893.

\bibitem{SILTP:CVPR:2010}
S.~Liao, G.~Zhao, V.~Kellokumpu, M.~Pietikäinen, S.~Z. Li, Modeling pixel
  process with scale invariant local patterns for background subtraction in
  complex scenes, in: CVPR, 2010, pp. 1301--1306.

\bibitem{Luo:ICCV:2013}
P.~Luo, X.~Wang, X.~Tang, Pedestrian parsing via deep decompositional network,
  in: ICCV, 2013, pp. 2648--2655.

\bibitem{Matsukawa:cvpr16}
T.~Matsukawa, T.~Okabe, E.~Suzuki, Y.~Sato, Hierarchical gaussian descriptor
  for person re-identification, in: 2016 IEEE Conference on Computer Vision and
  Pattern Recognition (CVPR), 2016, pp. 1363--1372.

\bibitem{Liao:CVPR:2015}
S.~Liao, Y.~Hu, X.~Zhu, S.~Z. Li, Person re-identification by local maximal
  occurrence representation and metric learning, in: CVPR, 2015, pp.
  2197--2206.

\bibitem{Mirmahboub:2017}
B.~Mirmahboub, M.~L. Mekhalfi, V.~Murino, Distance penalization and fusion for
  person re-identification, in: 2017 IEEE Winter Conference on Applications of
  Computer Vision (WACV), 2017, pp. 1306--1314.

\bibitem{Stuart:Science:2003}
J.~M. Stuart, E.~Segal, D.~Koller, S.~K. Kim, A gene-coexpression network for
  global discovery of conserved genetic modules, Science 302~(5643) (2003) 249
  -- 255.

\bibitem{Garcia:ICCV:2015}
J.~Garc\'ia, N.~Martinel, C.~Micheloni, A.~Gardel, Person re-identification
  ranking optimisation by discriminant context information analysis, in: 2015
  IEEE International Conference on Computer Vision (ICCV), 2015, pp.
  1305--1313.

\bibitem{Bai:CVPR17}
S.~{Bai}, X.~{Bai}, Q.~{Tian}, Scalable person re-identification on supervised
  smoothed manifold, in: CVPR, 2017, pp. 2530--2539.

\bibitem{Zhang:CVPR:2016}
Y.~Zhang, B.~Li, H.~Lu, A.~Irie, X.~Ruan, Sample-specific svm learning for
  person re-identification, in: CVPR, 2016, pp. 1278--1287.

\bibitem{Ahmed:CVPR:2015}
E.~Ahmed, M.~Jones, T.~K. Marks, An improved deep learning architecture for
  person re-identification, in: CVPR, 2015, pp. 3908--3916.

\bibitem{Liu:TIP17}
H.~Liu, J.~Feng, M.~Qi, J.~Jiang, S.~Yan, End-to-end comparative attention
  networks for person re-identification, IEEE Transactions on Image Processing
  26~(7) (2017) 3492--3506.

\bibitem{Du:ICPR:2012}
Y.~Du, H.~Ai, S.~Lao, Evaluation of color spaces for person re-identification,
  in: ICPR, 2012, pp. 1371--1374.

\bibitem{Gray:ECCV:2008}
D.~Gray, H.~Tao, Viewpoint invariant pedestrian recognition with an ensemble of
  localized features, in: ECCV, 2008, pp. 262--275.

\bibitem{Zheng:ICCV:2015}
L.~Zheng, L.~Shen, L.~Tian, S.~Wang, J.~Wang, Q.~Tian, Scalable person
  re-identification: A benchmark, in: ICCV, 2015.

\bibitem{Kodle:BIOINF:2012}
R.~Kolde, S.~Laur, P.~Adler, J.~Vilo, Robust rank aggregation for gene list
  integration and meta-analysis, Bioinformatics 28~(4) (2012) 573.

\bibitem{Chen:TPAMI17}
Y.~C. Chen, X.~Zhu, W.~S. Zheng, J.~H. Lai, Person re-identification by camera
  correlation aware feature augmentation, IEEE Transactions on Pattern Analysis
  and Machine Intelligence PP~(99) (2017) 1--14.

\bibitem{Zhao:CVPR17}
H.~Zhao, M.~Tian, S.~Sun, J.~Shao, J.~Yan, S.~Yi, X.~Wang, X.~Tang, Spindle
  net: Person re-identification with human body region guided feature
  decomposition and fusion, in: CVPR, 2017, pp. 1077--1085.

\bibitem{Jojic:CVPR:2009}
N.~Jojic, A.~Perina, M.~Cristani, V.~Murino, B.~Frey, Stel component analysis:
  Modeling spatial correlations in image class structure, in: CVPR, 2009, pp.
  2044--2051.

\end{thebibliography}

\end{document}